\documentclass{article}

\PassOptionsToPackage{numbers, compress}{natbib}
\usepackage[preprint]{nips_2018}

\usepackage[utf8]{inputenc} 
\usepackage[T1]{fontenc}    
\usepackage{url}            
\usepackage{booktabs}       
\usepackage{amsfonts}       
\usepackage{nicefrac}       
\usepackage{microtype}      

\usepackage{amsmath}
\usepackage{amssymb}
\usepackage{color}
\usepackage{epstopdf}
\usepackage{physics}
\usepackage{enumitem}
\usepackage{graphicx}

\newlist{step}{enumerate}{1}
\setlist[step]{label=Step-\arabic*:}

\date{18th May 2018}

\begin{document}


\title{Icing on the Cake: An Easy and Quick Post-Learnig Method You Can Try After Deep Learning}
\author{Tomohiko Konno\thanks{Corresponding Author: \color{blue}{tomohiko@nict.go.jp}}~~ and ~Michiaki Iwazume \vspace{2mm} \\  AI Science Research and Development Promotion Center \\National Institute of Information and Communications Technology, Tokyo Japan}
\maketitle
\begin{abstract}
We found an easy and quick post-learning method named ``Icing on the Cake'' to enhance a classification performance in deep learning. The method is that we train only the final classifier again after an ordinary training is done.
 \end{abstract}

\begin{figure}[h!]
\begin{center}
    \includegraphics[width=0.7\hsize]{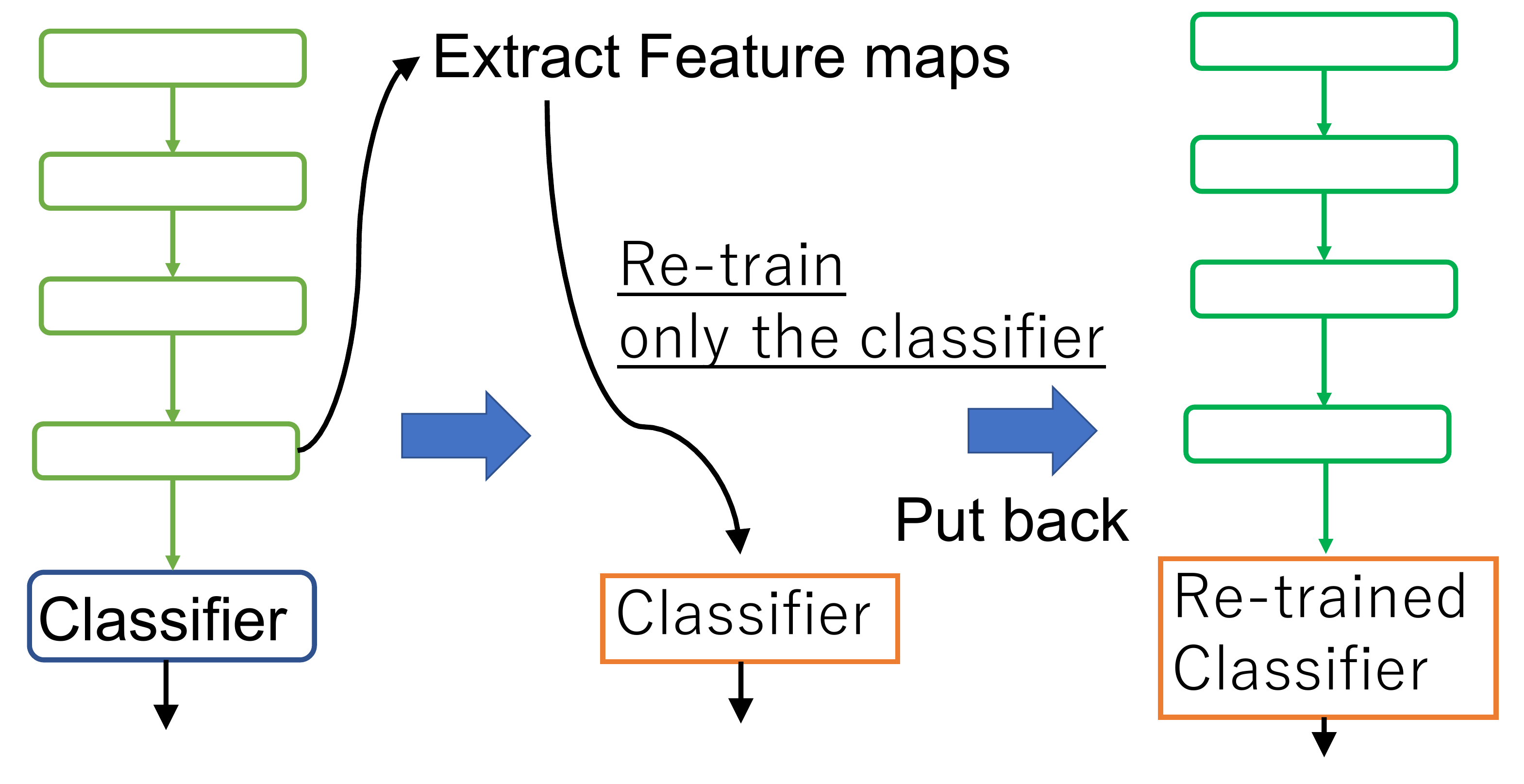}
    \caption{The sketch of the proposed method.(``Icing on the Cake''). Left: Train a deep neural network as usual. Center: Extract features of input data as estimation from the layer before the final classifier, and then train the final classifier again by the extracted features. Right: Put the re-trained classifier back to the network. The accuracies in cifar100 in ResNet56 is improved from 66\% to 73\% in our experiment }\label{fig: network_ice}
    \end{center}
\end{figure}

\section{Introduction and Related Works}
We have seen a great success in deep learning~\cite{goodfellow2016deep}. Now, deep leanings can do a lot of things including language translation, image classification, image generation~\cite{goodfellow2014generative,2018arXiv180401622J}, image generation from text~\cite{2018arXiv180105091H,2018arXiv180208216S,2018arXiv180209178Z,2018arXiv180308495C}, text generation from image~\cite{2018arXiv180400861L,2018arXiv180304376L}, unsupervised machine translation~\cite{2017arXiv171100043L,2017arXiv171011041A,2018arXiv180407755L}, playing Go (reinforcement learning)~\cite{silver2016mastering}, and even network design of deep learning by themselves~\cite{2018arXiv180203268P}, to raise a few. The achievements are so many, that we cannot exhaustively enough at all. 

Six years have passed since the advent of surprise to deep learning~\cite{krizhevsky2012imagenet}. However, there is no such symptom that the progress in deep learning research would slow-down at all. To the contrary, we see that the research has been accelerated, and the realm that deep learning achieves has been spreading even at this moment. We, human, cannot catch up the progress. We do not still understand deep learning yet. In the present paper, we will show another phenomena, which we found by chance.

Our proposed method named ``Icing on the Cake'' enhances the accuracies of image classification problems in deep learnings by an easy and quick way. We do not fully understand the reason yet, and the phenomena is interesting to the community.  It  is another learning method as: early stopping~\cite{prechelt1998early} that stops epochs early timing to prevent overfitting; dropout~\cite{srivastava2014dropout} that deactivates some nodes when training; batch normalization~\cite{ioffe2015batch};  group normalization~\cite{2018arXiv180308494W}; He initialization~\cite{2015arXiv150201852H}; Xavier initialization~\cite{glorot2010understanding}; mix-up (between-class learning; Sample Parings)~\cite{2017arXiv171009412Z, 2017arXiv171110284T,2018arXiv180102929I} that mixes datas and labels to enhance learning; and so forth. We add another one to the existing list.

\section{Proposed Method: Icing on the Cake}
The method named ``Icing on the Cake'' consists of what follows. The sketch of the idea is illustrated in Fig.~\ref{fig: network_ice}.
\begin{step}
\item Train a deep neural network as usual.
\item Extract features from the layer just before the final classifier as estimation.
\item Train only the final layer again by the extracted features.
\item Put the final classifier back to the original network.
\end{step}
After the procedure, accuracies increase.
\subsection{An Easy Way to Implement}
\begin{figure}[htbp]
\begin{center}
    \includegraphics[width=0.4\hsize]{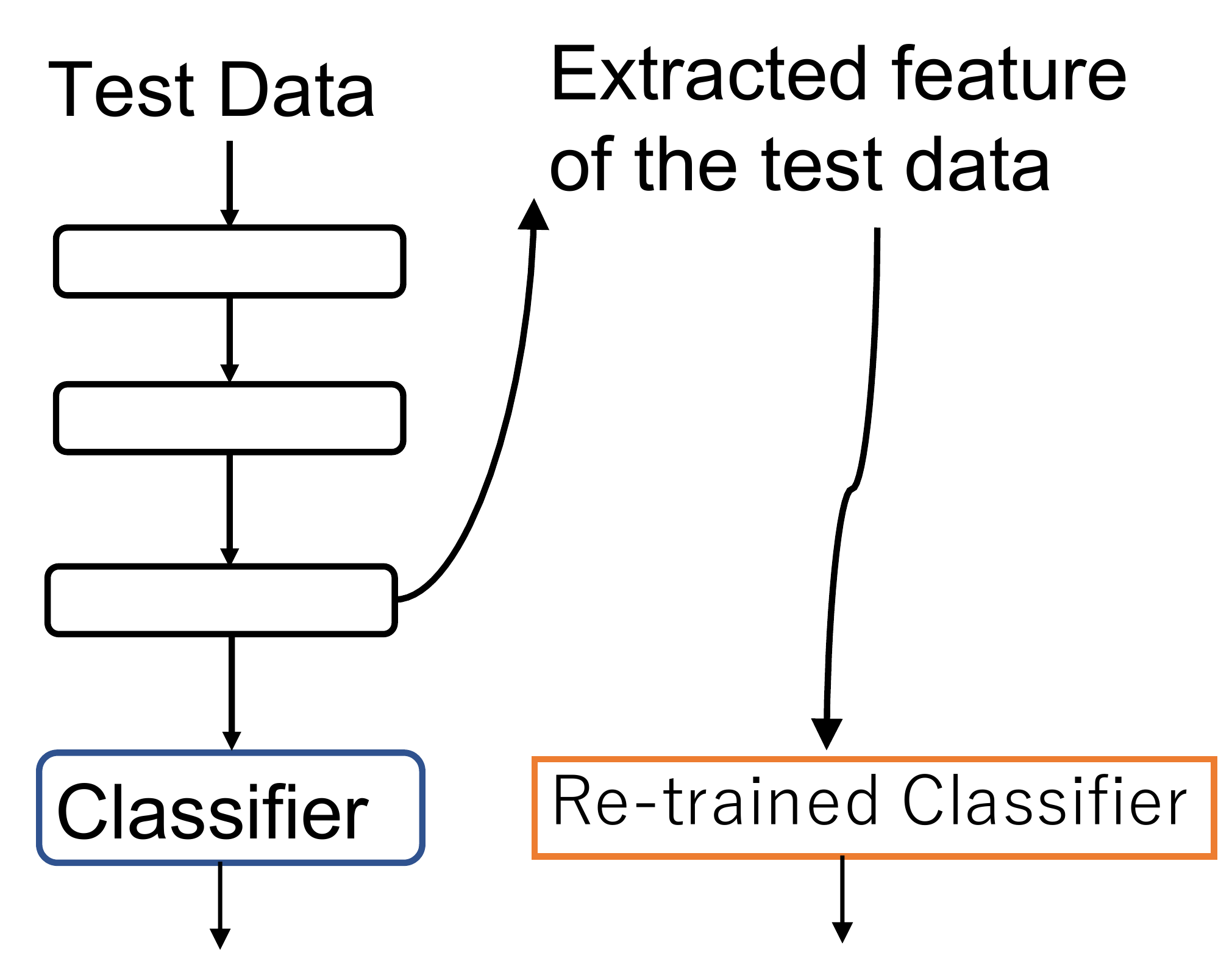}
    \caption{An easy way to implement when estimate}\label{fig: test_ice}
    \end{center}
\end{figure}

In the implementation, we do not  have to put the re-trained classifier back to the network. We have an easier equivalent method. First, we input test data, and then extract the features of the test data, and then input the features into the re-trained classifier. This is illustrated in Fig.~\ref{fig: test_ice}.

We have two good things in "Icing on the Cake".
\begin{enumerate}
\item It is easy to try.
\item It does not take time for re-training to finish.
\end{enumerate}
It is not very rare that it takes a week or so to train deep neural networks. However, since ``Icing on the Cake'' is just to train the final layer only, it does not take time. It is done in about five minutes or so. It is really icing on the cake. We do not strongly insist that ``Icing on the Cake'' works all the cases. We cannot prove it for all the networks to hold as of now. The method is worth trying, and it does not hurt.

\begin{figure}[h!]
\begin{center}
    \includegraphics[width=0.5\hsize]{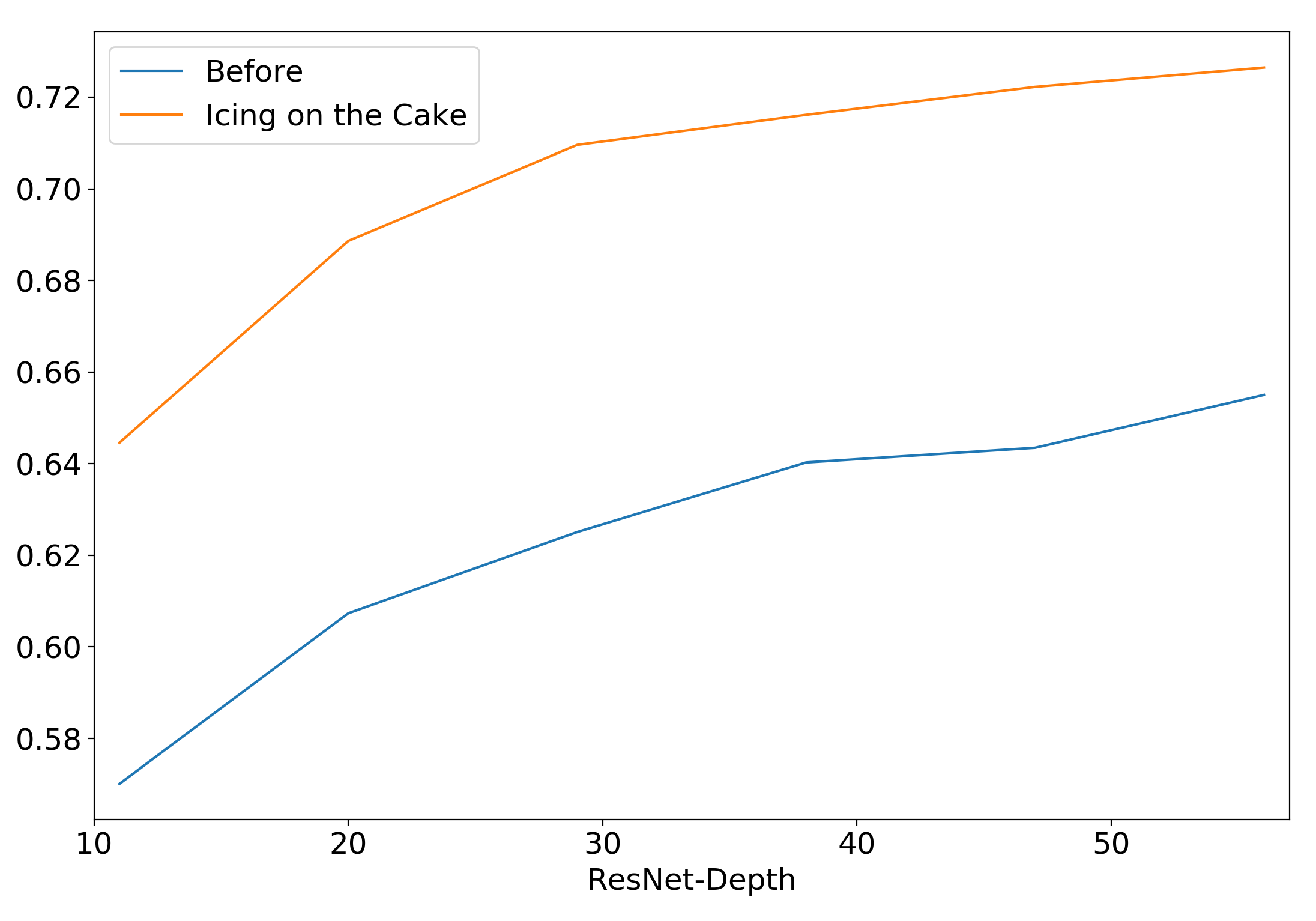}
    \caption{Accuracies in ResNet with varying depth from 11 to 56 for 100-image classification in cifar100 before and after ``Icing on the Cake''.}\label{fig: resnet}
    \end{center}
\end{figure}

\subsection{Experiments}
ResNet~\cite{he2015deep} with varying depth from 11 to 56 for the100-image classification problem in Cifar100~\cite{krizhevsky2009learning} are investigated.  Batch size is $128$, optimizer is Adam, epochs are $200$, and data augmentation is used. We use Keras~\cite{chollet2015keras}. The result is summarized in Tab.~\ref{tab: resnet-depth}, and illustrated in Fig.~\ref{fig: resnet}. The averages are taken over ten times. After "Icing on the Cake" the accuracies always get bigger than before, and the standard deviations get smaller.

In other experiments, ResNet56 and DenseNet\footnote{Depth=40, DenseBlock=3, and $k=12$}~\cite{DBLP:journals/corr/HuangLW16a} are investigated. The 100-image and 10-image classifications are done in Cifar100 and in Cifar10. The results are summarized in Tab.~\ref{tab: result100} and Tab.~\ref{tab: result10}. Batch size is $128$, optimizer is Adam, epochs are $200$, and data augmentation is used. "Icing on the Cake" is abbreviated as ICK in the tables. The results show that after "Icing on the Cake" the accuracies get clearly bigger than before. The results are the averages over ten data points for each. After ``Icing on the Cake" the accuracies get bigger, and the standard deviations get smaller.

\begin{table}[htbp]
    \begin{center}
        \begin{tabular}{|c||c|c|}\hline
  ResNet-Depth & Before &  Icing on the Cake \\ \hline
  11 &   0.57 (0.010)& 0.64 (0.003)\\ \hline
  20& 0.61 (0.014)&0.69 (0.005)    \\ \hline
  29 &0.63 (0.021)& 0.71 (0.005) \\ \hline
  38& 0.64 (0.016)&0.72 (0.004)  \\ \hline
  47& 0.64 (0.020)&0.72 (0.003) \\ \hline
  56& 0.66 (0.015)&0.73 (0.003) \\ \hline
\end{tabular}
\caption{The experiments in ResNet with varying depth from 11 to 56. Accuracies and stand deviations are written.}\label{tab: resnet-depth}
    \end{center}
    \end{table}

\begin{table}[htbp]
\begin{minipage}{0.5\hsize}
    \begin{center}
        \begin{tabular}{|c||c|c|c|c|}\hline
  &  ResNet56&  DenseNet \\ \hline
 
 Before&    0.66 (0.015)   & 0.59 (0.023)  \\ \hline
 ICK &   0.73 (0.003)&  0.79 (0.0049)    \\ \hline
\end{tabular}
\caption{The classifications in Cifar100}\label{tab: result100}
    \end{center}
    \end{minipage}
%
%
\begin{minipage}{0.5\hsize}
    \begin{center}
        \begin{tabular}{|c||c|c|c|c|}\hline
  &  ResNet56&  DenseNet \\ \hline
 Before&     0.885 (0.012)  & 0.86 (0.024)  \\ \hline
 ICK &  0.92 (0.0025)     &  0.91 (0.0057)    \\ \hline
\end{tabular}
\caption{The classifications in Cifar10}\label{tab: result10}
    \end{center}
            \end{minipage}
    \end{table}

\subsection{Comments}



We do not fully understand the reason why "Icing on the Cake" works. We did experiments in an effort to find a counter example. However, we could not. Rather, we collected more positive evidences at last. The reason why "Icing on the Cake" works is not yet fully investigated, and is open to the community.

\section{Conclusion}
We found that classification performance in deep learnings get improved with an easy and quick method named ``Icing on the Cake''. The method is just to train the final classifier only again by the extracted features. 

It is not that we strongly insist that every deep learning is improved by ``Icing on the Cake'', but that there is a possibility that it works well. It is very easy to try, and it does not take time. It is just to train the last one layer. It is worth trying. It is really icing on the cake.

\appendix

\section{Other experiments in WideResNet}
WideResNet--16--8~\cite{2016arXiv160507146Z} is investigated for 100-image classification problem in cifar100, and summarized in Tab.~\ref{tab: wideresnet}. The batch size is 128, optimizer is Adam, and the epochs are 200. The differences in Wide ResNets is not as big as in Resnet and DenseNet. However, after "Icing on the Cake", the accuracies get always bigger than before as in the table.

\begin{table}[htbp]
    \begin{center}
        \begin{tabular}{|c|c|}\hline
  Before&  Icing on the Cake \\ \hline
  0.72 & 0.73 \\ \hline
  0.72& 0.74     \\ \hline
  0.72&0.74 \\ \hline
  0.72& 0.73 \\ \hline
  0.72& 0.73\\ \hline
  0.72& 0.74 \\ \hline
  0.73&0.74 \\ \hline
  0.73& 0.74 \\ \hline
  0.72 & 0.73 \\ \hline
\end{tabular}
\caption{The experiments in WideResNet--16--8}\label{tab: wideresnet}
    \end{center}
\end{table}

\bibliographystyle{unsrt}
\bibliography{pseudo_data_generation.bib}
\end{document}